\documentclass[a4paper]{article}
\usepackage[utf8x]{inputenc}
\usepackage[T1]{fontenc}
\usepackage{import}

\usepackage[a4paper,margin=1in]{geometry}
\usepackage[colorlinks=true,allcolors=blue,pagebackref]{hyperref}
\usepackage[numbers]{natbib}

\title{A Limitation of the PAC-Bayes Framework}
\author{Roi Livni\thanks{Tel Aviv University, Department of Electrical Engineering} \and Shay Moran\thanks{Google AI, Princeton}}
\date{April 2020}

\usepackage[numbers]{natbib}
\usepackage{graphicx}
\usepackage{amsmath,amsthm,amsfonts}
\usepackage{xspace,hyperref,cleveref}
\usepackage{graphicx, color}
\newcommand{\pos}{\mathsf{pos}}
\newcommand{\alg}{\mathcal{A}}
\newcommand{\growth}{\Phi}

\usepackage{hyperref}
\newcommand{\N}{\mathbb{N}}

\newtheorem{property}{Property}
\newcommand{\ignore}[1]{}
\newcommand{\twr}{\mathbf{twr}}
\newtheorem{theorem}{Theorem}
\newtheorem{corollary}{Corollary}
\newtheorem{definition}{Definition}
\newtheorem{lemma}{Lemma}
\newtheorem{claim}{Claim}

\newcommand{\E}{\mathop\mathbb{E}}
\newcommand{\X}{\mathcal{X}}
\renewcommand{\H}{\mathcal{H}}
\newcommand{\Y}{\mathcal{Y}}

\newcommand{\Loss}{\mathcal{L}}

\newcommand{\vv}{\mathbf{h}}
\newcommand{\hh}{\mathbf{h}}

\newcommand{\kl}[2]{\mathrm{KL}\left(#1\|#2\right)}

\newcommand{\new}[1]{{\color{black} #1}}

\renewcommand{\alg}{\mathcal{A}}
\Crefname{claim}{Claim}{Claims}
\newcommand{\fullversion}[1]{the full version  \citep{fullversion}}
\begin{document}

\maketitle
\begin{abstract}
PAC-Bayes is a useful framework for deriving generalization bounds which was introduced by McAllester ('98).
	This framework has the flexibility of deriving distribution- and algorithm-dependent bounds, 
	which are often tighter than VC-related uniform convergence bounds. In this manuscript we present a limitation for the PAC-Bayes framework.
	We demonstrate an easy learning task which is not amenable to a PAC-Bayes analysis.

Specifically, we consider the task of linear classification in 1D;
	it is well-known that this task is learnable using just $O(\log(1/\delta)/\epsilon)$ examples. 
%	and that this sample complexity is achieved by any Empirical Risk Minimizer.
	On the other hand, we show that this fact can not be proved using a PAC-Bayes analysis: 
	for any algorithm that learns 1-dimensional linear classifiers 
	there exists a (realizable) distribution for which the PAC-Bayes bound is arbitrarily large.

\end{abstract}
\section{Introduction}

The classical setting of supervised binary classification considers \emph{learning algorithms} that receive (binary) labelled examples and are required to output a \emph{predictor} or a \emph{classifier} that predicts the label of new and unseen examples. Within this setting, Probably Approximately Correct (PAC) generalization bounds quantify the success of an algorithm to approximately predict with high probability. The PAC-Bayes framework, introduced in \cite{mcallester1999some,shawe1997pac} and further developed in \cite{mcallester1999pac,mcallester2003simplified,seeger2002pac}, provides PAC-flavored bounds to Bayesian algorithms that produce \emph{Gibbs-classifiers} (also called \emph{stochastic-classifiers}). These are classifiers that, instead of outputting a single classifier, output a probability distribution over the family of classifiers. Their performance is measured by the expected success of prediction where expectation is taken with respect to both sampled data and sampled classifier.

A PAC-Bayes generalization bound relates the generalization error of the algorithm to a KL distance between the stochastic output classifier and some \emph{prior distribution} $P$. In more detail, the generalization bound is comprised of two terms: first, the empirical error of the output Gibbs-classifier, and second, the KL distance between the output Gibbs classifier and some arbitrary (but sample-independent) prior distribution. This standard bound captures a basic intuition that a good learner needs to balance between bias, manifested in the form of a prior, and fitting the data, which is measured by the empirical loss.
A natural task is then, to try and characterize the potential as well as limitations of such Gibbs-learners that are amenable to PAC-Bayes analysis. As far as the potential, several past results established the strength and utility of this framework (e.g. \cite{shawe2009pac, seeger2001improved, langford2003pac, dziugaite2017computing, langford2002not}).

In this work we focus on the complementary task, and present the first limitation result showing that there are classes that are learnable, even in the strong distribution-independent setting of PAC, but do not admit any algorithm that is amenable to a non-vacuous PAC-Bayes analysis. We stress that this is true even if we exploit the bound to its fullest and allow any algorithm and any possible, potentially distribution-dependent, prior.

More concretely, we consider the class of 1-dimensional thresholds, i.e.\ the class of linear classifiers over the real line.
It is a well known fact that this class is learnable and enjoys highly optimistic sample complexity.
Perhaps surprisingly, though, we show that any Gibbs-classifier that learns the class of thresholds, must output posteriors from an unbounded set. We emphasize that the result is provided even for priors that depend on the data distribution. %\new{As such, PAC-Bayes bounds are considered a powerful tool in the exploration of \emph{distribution-dependent} generalization bounds \cite{dziugaite2018data, lever2010distribution, parrado2012pac, rivasplata2018pac}, and our result, then, is given in this very general setting.}
%

%The above result is achieved over the infinite class of thresholds. For the class of thresholds over a finite set, say of integers, we can show that the PAC-Bayes bound depends on the \emph{domain-size}. That is true even though, in general, the class is learnable with sample complexity that is independent of the domain-size. In other words, by choosing a class of thresholds over sufficiently large domain, we can show an arbitrary gap between the true sample complexity of the problem and the one obtained via a PAC-Bayes bound.

From a technical perspective our proof exploits and expands a technique that was recently introduced by Alon et al. \cite{alon} to establish limitations on differentially-private PAC learning algorithms. The argument here follow similar lines, and we believe that these similarities in fact highlight a potentially powerful method to derive further limitation results, especially in the context of stability.

\section{Preliminaries}
\subsection{Problem Setup}
We consider the standard setting of binary classification.  
Let  $\X$ denote the domain and $\Y=\{\pm1\}$ the label space.
We study learning algorithms that observe as input a sample $S$ of labelled examples 
drawn independently from an unknown target distribution $D$, supported on $\X\times \Y$.
The output of the algorithm is an hypothesis $h:\X\to \Y$,
and its goal is to minimize the $0/1$-loss, which is defined by:
\[\Loss_D(h)= \E_{(x,y)\sim D}\bigl[{\bf 1}[h(x)\neq y]\bigr].\]
We will focus on the setting where the distribution $D$ is \emph{realizable} with respect to a fixed hypothesis class $\H\subseteq \Y^\X$ which is known in advance. 
That is, it is assumed that there exists $h\in \H$ such that:
$\Loss_D(h)=0$. Let $S=\langle (x_1,y_1),\ldots, (x_m,y_m)\rangle \in (\X\times \Y)^m$ be a sample of labelled examples. 
	The empirical error $\Loss_S$ with respect to $S$ is defined by
\[\Loss_{S}(h)= \frac{1}{m}\sum_{i=1}^m \mathbf{1}[h(x)\ne y].\]
\new{We will use the following notation:
for a sample $S=\langle (x_1,y_1),\ldots (x_m,y_m)\rangle$, let $\underline{S}$ denote
the underlying set of unlabeled examples $\underline{S} = \{x_i : i\leq m\}$.}
%The sequence $\langle y_1,\ldots, y_m\rangle$ is called the \emph{sign-pattern} of $S$. 

\paragraph{The Class of Thresholds.}
For $k\in\N$ let $h_k:\N\to\{\pm 1\}$ denote the {\it threshold function}
\[
h_{k}(x)=\begin{cases}
-1 & x\le k\\
{+}1 & x>k.
\end{cases}
\]

The class of thresholds $\H_\N$ is the class $\H_\N:=\{h_k : k\in\N\}$ over the domain $\X_\N:=\N$.
Similarly, for a finite $n\in \N$ let $\H_n$ denote the class of all thresholds restricted to the domain $\X_n:=[n]=\{1,\ldots, n\}$.
%A sample $S=\langle (x_1,y_1),\ldots, (x_m,y_m)\rangle \in (\X\times \Y)^m$ is said to be {\it realizable} by a threhsold,
%if there exists a threshold $h_k$ such that $h_k(x_i)=y_i$ for all $i$.
Note that $S$ is realizable with respect to $\H_\N$ if and only if either (i) $y_i=+1$ for all $i\leq m$, or (ii) there exists $1\le j\le m$ such that $y_i=-1$ if and only if $x_i\le x_j$. %We will call $j$ the {\it threshold} of the sign-pattern.

\vspace{1mm}

A basic fact in statistical learning is that $\H_\N$ is PAC-learnable.
	That is, there exists an algorithm $A$ such that for every realizable distribution $D$, if $A$ is given a sample of size $O(\frac{\log 1/\delta}{\epsilon})$ examples drawn from $D$,
	then with probability at least $1-\delta$, the output hypothesis $h_S$ satisfies $\Loss_D(h_{S})\le \epsilon$.
	In fact, any algorithm $A$ which returns an hypothesis $h_k\in \H_\N$ which is consistent with the input sample,
	will satisfy the above guarantee.
	Such algorithms are called empirical risk minimizers (ERMs).
	We stress that the above sample complexity bound is \emph{independent} of the domain size. 
	In particular it applies to $\H_n$ for every $n$, as well as to the infinite class $\H_\N$.
	For further reading, we refer to text books on the subject, such as \cite{shai, mohri}.
%
%
%In fact, the algorithm $A$ is extremely simple: Given a sample $S$ we choose $h\in \H_n$ that minimizes the empirical error, namely $\Loss_{S}(h)=0$. The most basic facts in learning theory demonstrate that this algorithm (termed \emph{empirical risk minimizer} (ERM)) succeeds w.h.p. We refer the reader to standard text books for further reading \cite{shai, mohri}.

\subsection{PAC-Bayes Bounds}\label{sec:pacbayes}
PAC Bayes bounds are concerned with \emph{stochastic-classifiers}, or \emph{Gibbs-classifiers}. 
	A Gibbs-classifier is defined by a distribution $Q$ over hypotheses.
	The distribution $Q$ is sometimes referred to as a \emph{posterior}.
	The loss of a Gibbs-classifier with respect to a distribution $D$ is given by the expected loss over the drawn hypothesis and test point, namely: \[\Loss_D(Q)= \E_{h\sim Q, (x,y)\sim D}[{\bf 1}\bigl[h(x)\neq y]\bigr].\]
%	Note that every deterministic hypothesis $h$ can be seen as a distribution $Q_h$ supported only on $h$,	and therefore Gibbs-classifiers are strictly more general than (deterministic) hypotheses.

A key advantage of the PAC-Bayes framework is its flexibility of deriving generalization bounds that do not depend on an hypothesis class. Instead, they provide bounds that depend on the KL distance between the output posterior and a fixed prior $P$.  
	Recall that the KL divergence between a distribution $P$ and a distribution $Q$ is defined as follows\footnote{We use here the standard convention that if $P(\{x: Q(x)=0\})>0$  then $\kl{P}{Q}=\infty$.}:
	\[\kl{P}{Q}=\E_{x\sim P}\Bigl[\log \frac{P(x)}{Q(x)}\Bigr].\]
	
Then, the classical PAC-Bayes bound asserts the following:
\begin{theorem}[PAC-Bayes Generalization Bound \citep{mcallester1999some}]\label{thm:pacbayes}
Let $D$ be a distribution over examples, let $P$ be a \emph{prior distribution} over hypothesis, and let $\delta>0$.
Denote by $S$ a sample of size $m$ drawn independently from $D$.
Then, the following event occurs with probability at least $1-\delta$:
for every \emph{posterior distribution}~$Q$,
\begin{align*}
\Loss_D(Q)\le \Loss_{S}(Q) + O\left( \sqrt{\frac{\kl{Q}{P}+ \ln \sqrt{m}/\delta}{m}}\right).
\end{align*}
\end{theorem}

The above bound relates the generalization error to the KL divergence between the posterior and the prior. 
Remarkably, the prior distribution $P$ can be chosen as a function of the target distribution $D$, allowing to obtain distribution-dependent generalization bounds.

 Since this pioneer work of \citet{mcallester1999pac}, many variations on the PAC-Bayes bounds have been proposed. Notably, \citet{seeger2001improved} and \citet{catoni2007pac} provided bounds that are known to converge at rate $1/m$ in the realizable case (see also \cite{guedj2019primer} for an up-to-date survey). We note that our constructions are all provided in the realizable setting, hence readily apply.

\section{Main Result}
We next present the main result in this manuscript. Proofs are provided in \Cref{sec:prfs}.  
The statements use the following function $\growth(m,\gamma,n)$, which is defined for $m,n>1$ and $\gamma\in (0,1)$:
\begin{align*}
\growth(m,\gamma,n)=
\frac{\log^{(m)}(n)}{(\frac{10m}{\gamma})^{3m} }.
\end{align*}
Here, $\log^{(k)}(x)$ denotes the iterated logarithm, i.e.\ 
\[\log^{(k)}(x) = \underbrace{\log(\log\ldots (\log(x)) )}_{k \text{ times}}.\]
An important observation is that $\lim_{n\to \infty}\growth(m,\gamma,n)=\infty$ for every fixed $m$ and~$\gamma$.

% which is defined by the recursion% \[% \log^{(k)}(m) = % \begin{cases}% m &k=0,\\% \log(\log^{(k-1}(m)) &k>0.% \end{cases}% \]
\begin{theorem}[Main Result]\label{thm:main}
Let $n,m > 1$ be integers, and let $\gamma\in (0,1)$. 
Consider the class $\H_n$ of thresholds over the domain $\X_n=[n]$. 
Then, for any learning algorithm $A$ which is defined on samples of size $m$, 
there exists a realizable distribution $D=D_A$ such that for any prior $P$
the following event occurs with probability at least $1/16$ over the input sample $S\sim D^m$,
\[\kl{Q_S}{P} =\tilde \Omega\left(\frac{\gamma ^2}{m^2}\log\Bigl(\new{\frac{\growth(m,\gamma,n)}{m}}\Bigr)\right)
\quad \mathrm{or} \quad \Loss_D(Q_{S})> 1/2-\gamma-\frac{m}{\growth(m,\gamma,n)},\]
where $Q_{S}$ denotes the posterior outputted by~$A$.
\end{theorem}
To demonstrate how this result implies a limitation of the PAC-Bayes framework, 
    pick $\gamma=1/4$ and consider any algorithm $A$ which learns thresholds over the natural numbers $\X_\N=\N$ with confidence $1-\delta\geq 99/100$, error $\epsilon< 1/2-\gamma=1/4$, and $m$ examples\footnote{We note in passing that any Empirical Risk Minimizer learns thresholds with these parameters using $<50$ examples.}.  
	Since $\growth(m,1/4,n)$ tends to infinity with $n$ for any fixed $m$, the above result implies the existence of a realizable distribution $D_n$ supported on $X_n\subseteq\N$ such that the PAC-Bayes bound with respect to any possible prior $P$  
	will produce vacuous bounds. We summarize it in the following corollary.
\begin{corollary}[PAC-learnability of Linear classifiers cannot be explained by PAC-Bayes]\label{thm:main2}
Let $\H_\N$ denote the class of thresholds over $\X_\N=\N$ and let $m>0$.
Then, for every algorithm $A$ that maps inputs sample $S$ of size $m$ to output posteriors $Q_S$
and for every arbitrarily large $N>0$ there exists a realizable distribution $D$ 
such that, for any prior $P$, with probability at least $1/16$ over $S\sim D^m$ on of the following holds:
\[
\kl{Q_S}{P} > N
\qquad \mathrm{or}, \qquad
\Loss_D(Q_{S})>1/4.\] 
\end{corollary}

A different interpretation of \Cref{thm:main} is that 
    in order to derive meaningful PAC-Bayes generalization bounds for PAC-learning thresholds over a finite domain $X_n$,
	the sample complexity must grow to infinity with the domain size~$n$ (it is at least $\Omega(\log^\star(n)$)).
	In contrast, the true sample complexity of this problem is $O(\log (1/\delta)/\epsilon)$ which is independent of $n$.

  \section{Technical Overview}
 
 A common approach of proving impossibility results in computer science (and in machine learning in particular)
	 exploits a Minmax principle, whereby one specifies a fixed hard distribution over inputs,
	 and establishes the desired impossibility result for any algorithm with respect to random inputs from that distribution.
	As an example, consider the ``No-Free-Lunch Theorem''
	which establishes that the VC dimension lower bounds the sample complexity of PAC-learning a class~$\H$. 
	Here, one fixes the distribution to be uniform over a shattered set of size $d=\mathsf{VC}(H)$,
	and argues that every learning algorithm must observe $\Omega(d)$ examples. (See e.g.\ Theorem~5.1 in \cite{shai}.)
 
Such ``Minmax'' proofs establish a stronger assertion: they apply even to algorithms that ``know'' the input-distribution.
	For example, the No-Free-Lunch Theorem applies even to learning algorithms that are designed
	given the knowledge that the marginal distribution is uniform over some shattered set.

Interestingly, such an approach is bound to fail in proving \Cref{thm:main}.
	The reason is that if the marginal distribution $D_\X$ over $\X_n$ is fixed,
	then one can pick an $\epsilon/2$-cover\footnote{I.e.\ $\mathcal{C}_n$ satisfies that $(\forall h\in \H_n)(\exists c\in \mathcal{C}_n): \Pr_{x\sim D_X}(c(x)\neq h(x))\leq\epsilon/2$.} $\mathcal{C}_n\subseteq \H_n$ 
	of size $\lvert \mathcal{C}_n\rvert = O(1/\epsilon)$,
	and use any Empirical Risk Minimizer for $\mathcal{C}_n$.
	Then, by picking the prior distribution~$P$ to be uniform over $\mathcal{C}_n$,
	one obtains a PAC-Bayes bound which scales with the entropy $H(P)=\log\lvert \mathcal{C}_n\rvert=O(\log(1/\epsilon))$,
	and yields a $\mathsf{poly}(1/\epsilon,\log(1/\delta))$ generalization bound, which is independent of $n$.
	In other words, in the context of \Cref{thm:main}, there is no single distribution which is ``hard'' for all algorithms.

Thus, to overcome this difficulty one must come up with a ``method''
	which assigns to any given algorithm $A$ a ``hard''  distribution $D=D_A$, which witnesses \Cref{thm:main} with respect to $A$.
	The challenge is that $A$ is an arbitrary algorithm;
	e.g.\ it may be improper\footnote{I.e.\ $A$ may output hypotheses which are not thresholds, or Gibbs-classifiers supported on hypotheses which are not thresholds. }
	or add different sorts of noise to its output classifier.
	\new{We refer the reader to~\cite{NachumSY18,NachumY19,BassilyMNSY18} 
    for a line of work which explores in detail a similar ``failure'' of
    the Minmax principle in the context of PAC learning with low mutual information.}

The method we use in the proof of \Cref{thm:main} exploits Ramsey Theory.
	In a nutshell, Ramsey Theory provides powerful tools
	which allow to detect, for any learning algorithm, 
	a large \emph{homogeneous} set 
	such that the behavior of $A$ on inputs from the homogeneous set is highly regular.
	Then, we consider the uniform distribution over the homogeneous set
	to establish \Cref{thm:main}.

We note that similar applications of Ramsey Theory in proving lower bounds in computer science
	date back to the 80's~\citep{moran1985applications}.
	For more recent usages see e.g.~\citep{Bun16thesis,Correa19prophet,Cohen19Screen,alon}.
	Our proof closely follows the argument of \citet{alon},
	which establishes an impossibility result for learning $\H_n$ by differentially-private algorithms.

\paragraph{Technical Comparison with the Work by Alon et al.~\cite{alon}.}	
For readers who are familiar with the work of \cite{alon}, 
	let us summarize the main differences between the two proofs.
	The main challenge in extending the technique from \cite{alon} to prove \Cref{thm:main}  
	is that PAC-Bayes bounds are only required to hold for \emph{typical samples}.
	This is unlike the notion of differential-privacy (which was the focus of \cite{alon}) 
	that is defined with respect to \emph{all samples}. 
	Thus, establishing a lower bound in the context of differential privacy is easier: 
	one only needs to demonstrate a single sample for which privacy is breached. 
	However, to prove \Cref{thm:main} one has to demonstrate that the lower bound applies to many samples.
	Concretely, this affects the following parts of the proof:
	\begin{itemize}
	\item[(i)] The Ramsey argument in the current manuscript (\Cref{lem:homo}) is more complex:
	to overcome the above difficulty we needed to modify the coloring and the overall construction is more convoluted.
	\item[(ii)] Once Ramsey Theorem is applied and the homogeneous subset $R_n\subseteq X_n$ is derived,
	one still needs to derive a lower bound on the PAC-Bayes quantity.
	This requires a technical argument (\Cref{lem:sensitivity}), which is tailored to the definition of PAC-Bayes.
	Again, this lemma is more complicated than the corresponding lemma in \cite{alon}.
	\item[(iii)] Even with \Cref{lem:homo} and \Cref{lem:sensitivity} in hand, 
	the remaining derivation of \Cref{thm:main} still requires a careful analysis 
	which involves defining several ``bad'' events and bounding their probabilities.
	Again, this is all a consequence of that the PAC-Bayes quantity is an ``average-case'' complexity measure.
	\end{itemize}
\subsection{Proof Sketch and Key Definitions}
The proof of \Cref{thm:main} consists of two steps: 
    (i) detecting a hard distribution $D=D_A$ which witnesses
	\Cref{thm:main} with respect to the assumed algorithm $A$, and 
	(ii) establishing the conclusion of \Cref{thm:main} given the hard distribution $D$.
	The first part is combinatorial (exploits Ramsey Theory), 
	and the second part is more information-theoretic. 
	For the purpose of exposition, we focus in this technical overview, on a specific algorithm $\alg$. 
	This will make the introduction of the key definitions and presentation of the main technical tools more accessible.

\paragraph{The algorithm $\alg$.}
Let $S=\langle (x_1,y_1),\ldots, (x_m,y_m)\rangle$ be an input sample.
The algorithm $\alg$ outputs the posterior distribution~$Q_S$ which is defined as follows:
let $h_{x_i}={\bf 1}[x > x_i] - {\bf 1}[x\leq x_i]$ denote the threshold corresponding to the $i$'th input example.
The posterior $Q_S$ is supported on $\{h_{x_i}\}_{i=1}^m$, 
and to each $h_{x_i}$ it assigns a probability according to a decreasing function of its empirical risk.
(So, hypotheses with lower risk are more probable.)
The specific choice of the decreasing function does not matter, 
but for concreteness let us pick the function $\exp(-x)$. Thus,
\begin{equation}\label{eq:alg_example}
Q_S(h_{x_i})\propto \exp\bigl(-\Loss_{S}(h_{x_i})\bigr).
\end{equation} 
While one can directly prove that the above algorithm does not admit a PAC-Bayes analysis, 
we provide here an argument which follows the lines of the general case.
We start by explaining the key property of \emph{Homogeneity}, which allows to detect the hard distribution.

\subsubsection{Detecting a Hard Distribution: {Homogeneity}} 
The first step in the proof of \Cref{thm:main} takes the given algorithm 
and identifies a large subset of the domain on which its behavior is \emph{Homogeneous}.
In particular, we will soon see that the algorithm~$\alg$ is \emph{Homogeneous} on the entire domain~$\X_n$.
In order to define Homogeneity, we use the following equivalence relation between samples:
\begin{definition}[Equivalent Samples]\label{def:equivalent_samples}
Let $S=\langle (x_1,y_1),\ldots, (x_m,y_m)\rangle$ and $S'=\langle (x_1',y_1'),\ldots, (x_m',y_m')\rangle$
	be two samples.
	We say that $S$ and $S'$ are equivalent if for all $i,j\leq m$ the following holds.
	\begin{enumerate}
	\item $x_i \leq x_j\iff x_i'\leq x_j'$, and
	\item $y_i=y_i'$.
	\end{enumerate}
\end{definition}
For example, $\langle(1,-),(5,+),(8,+)\rangle$ and $\langle(10,-),(70,+),(100,+)\rangle$ are equivalent, 
but $\langle(3,-), (6,+), (4,+)\rangle$ is not equivalent to them (because of Item 1). 
For a point $x\in \X_n$ let $\pos(x;S)$ denote the number of examples in $S$ that are less than or equal to $x$:
\begin{align}\label{eq:pos}
\pos(x;S)= \Bigl\lvert\{x_i\in\underline{S} : x_i\le x\}\Bigr\rvert.
\end{align}
For a sample $S=\langle (x_1,y_1),\ldots, (x_m,y_m)\rangle$ let $\pi(S)$ denote the \emph{order-type} of $S$:
\begin{align}\label{eq:order-type}
\pi(S) = (\pos(x_1;S), \pos(x_2;S),\ldots, \pos(x_m;S)).
\end{align}
So, the samples $\langle(1,-),(5,+),(8,+)\rangle$ and $\langle(10,-),(70,+),(100,+)\rangle$ have order-type $\pi=(1,2,3)$,
whereas $\langle(3,-), (6,+), (4,+)\rangle$ has order-type $\pi=(1,3,2)$.

Note that $S,S'$ are equivalent if and only if they have the same labels-vectors and the same order-type.
	Thus, we encode the equivalence class of a sample by the pair $(\pi, \bar y)$, 
	where $\pi$ denotes its order-type and $\bar y=(y_1\ldots y_m)$ denotes its labels-vector. 
	The pair $(\pi,y)$ is called the \emph{equivalence-type of $S$.}

We claim that $\alg$ satisfies the following property of \emph{Homogeneity}:
\begin{property}[Homogeneity]\label{def:homo}
The algorithm $\alg$ possesses the following property:
	for every two equivalent samples $S,S'$
	and every $x,x'\in \X_n$ such that $\pos(x,S)=\pos(x',S')$,
\[\Pr_{h\sim Q_S}[h(x)=1]=\Pr_{h'\sim Q_{S'}}[h'(x')=1],\]
	where $Q_S,Q_{S'}$ denote the Gibbs-classifier outputted by $\alg$ on the samples $S,S'$.
\end{property}
In short, Homogeneity means that the probability $h\sim Q_S$ satisfies $h(x)=1$
	depends only on $\pos(x,S)$ and on the equivalence-type of~$S$.
    To see that $\alg$ is indeed homogeneous, let $S,S'$ be equivalent samples 
	and let $Q_S, Q_{S'}$ denote the corresponding Gibbs-classifiers outputted by $\alg$. 
	Then, for every $x,x'$ such that $\pos(x,S)=\pos(x',S')$, \Cref{eq:alg_example} yields that:
	\[\Pr_{h\sim Q_S}\bigl[h(x)=+1\bigr] = \sum_{x_i < x}Q_S(h_{x_i}) = \sum_{x_i' < x'}Q_{S'}(h_{x_i'}) = \Pr_{h'\sim Q_{S'}}\bigl[h'(x')=+1\bigr],\]
	where in the second transition we used that $Q_S(h_{x_i})= Q_{S'}(h_{x_i'})$ for every $i\leq m$ (because $S,S'$ are equivalent),
	and that $x_i \leq x \iff x_i'\leq x'$, for every $i$ (because $\pos(x,S)=\pos(x',S')$).

\paragraph{The General Case: Approximate Homogeneity.}
Before we continue to define the hard distribution for algorithm $A$, 
	let us discuss how the proof of \Cref{thm:main} handles arbitrary algorithms that are not necessarily homogeneous.

The general case complicates the argument in two ways.
	First, the notion of Homogeneity is relaxed to an approximate variant which is defined next.
	Here, an order type $\pi$ is called a \emph{permutation} if $\pi(i)\neq\pi(j)$ for every distinct $i,j\leq m$.
	(Indeed, in this case $\pi=(\pi(x_1)\ldots \pi(x_m))$ is a permutation of $1\ldots m$.)
	Note that the order type of $S=\langle(x_1,y_1)\ldots (x_m,y_m))\rangle$ is a permutation
	if and only if all the points in $S$ are distinct (i.e.\ $x_i\neq x_j$ for all $i\neq j$).

\begin{definition}[Approximate Homogeneity]\label{def:approxhomo}
An algorithm $\cal{B}$ is $\gamma$-approximately $m$-homogeneous if the following holds:
	let $S,S'$ be two equivalent samples of length $m$ whose order-type is a permutation,
	and let $x\notin \underline{S},x'\notin \underline{S}'$ such that $\pos(x,S)=\pos(x',S')$. Then,
 \begin{equation}\label{eq:approxhomo}
\lvert Q_S(x)-Q_{S'}(x')\rvert \le \frac{\gamma}{5m},
 \end{equation}
	where $Q_S,Q_{S'}$ denote the Gibbs-classifier outputted by $\mathcal{B}$ on the samples $S,S'$.
\end{definition}
%\begin{remark} 
%The above definition is restricted only to samples whose order-type is a permutation (i.e.\ whose all points are distinct).
%One could also consider a variant of homogeneity which applies to all samples of all order-types,
%While this may have been more natural, we chose the more restricted definition as it leads
%to much better quantitative bounds on the size of the approximate homogenous set.
%\end{remark}
Second, we need to identify a sufficiently large subdomain on which the assumed algorithm is approximately homogeneous.
	This is achieved by the next lemma, which is based on a Ramsey argument.
\begin{lemma}[Large Approximately Homogeneous Sets ]\label{lem:homo}
Let $m,n>1$ and let $\mathcal{B}$ be an algorithm that is defined over input samples of size $m$ over $\X_n$. 
Then, there is $\X'\subseteq \X_n$ of size $\lvert \X'\rvert\ge \growth(m,\gamma,n)$
such that the restriction of $\cal{B}$ to input samples from $\X'$ is $\gamma$-approximate $m$-homogeneous.
\end{lemma}
%\Cref{lem:homo} is a slight strengthening of Lemma 9 in \cite{alon}. 
We prove \Cref{lem:homo} in  \Cref{prf:homo}.
	For the rest of this exposition we rely on \Cref{def:homo} as it simplifies the presentation of the main ideas.

\paragraph{The Hard Distribution $D$.}
We are now ready to finish the first step and define the ``hard'' distribution~$D$.	
Define $D$ to be uniform over examples $(x,y)$ such that $y=h_{n/2}(x)$.
	So, each drawn example $(x,y)$ satisfies that $x$ is uniform in $\X_n$ and $y=-1$ if and only if $x\leq n/2$.
	In the general case,~$D$ will be defined in the same way with respect to the detected homogeneous subdomain.

\subsubsection{Hard Distribution $\implies$ Lower Bound: Sensitivity }

We next outline the second step of the proof, which establishes \Cref{thm:main} using the hard distribution~$D$.
Specifically, we show that for a sample~$S\sim D^m$,
\[\kl{Q_S}{P} =\tilde \Omega\left({\frac{1 }{m^2}}\log(\lvert \X_n\rvert)\right),\]
with a constant probability bounded away from zero. (In the general case $\lvert X_n\rvert$ is replaced by
$\growth(m,\gamma,n)$ -- the size of the homogeneous set.)

%To achieve this, it will be convenient to imagine that $S$ is drawn in two steps,
%	where in step (i) an equivalence type $(\pi,y)$ is sampled and in step (ii) $S$ is sampled conditioned on having type $(\pi,y)$.
%	To help digesting this process (and recalling the definition of a type), 
%	consider for example the type $(\pi,y)$, where $\pi=(1,2,\ldots,m)$ and $y=(+,+,\ldots,+)$.
%	In other words, the $x_i$'s satisfy $x_1<x_2<\ldots<x_m$ and the $y_i$'s are all $+1$.
%	Thus, $S$ has type $(\pi,y)$ if and only if $n/2\leq x_1 < x_2 <\ldots x_m$, 
%	which happens with probability 
%	\[\Pr_{S\sim D^m}\bigl[S \mbox{ has type $(\pi,y)$}\bigr]=\frac{{n/2 \choose m}}{n^m}.\]
\paragraph{Sensitive Indices.}
We begin with describing the key property of homogeneous learners.
Let $(\pi, \bar y)$ denote the equivalence-type of the input sample $S$.
	By homogeneity (\Cref{def:homo}), there is a list of numbers $p_0,\ldots, p_{m}$, 
	which depends only on the order-type $(\pi, \bar y)$, such that
	$\Pr_{h\sim Q_S}[h(x)=1] = p_i$
	for every $x\in\X_n$, where $i=\pos(x,S)$.
	The crucial observation is that there exists an index $i\leq m'$ which is \emph{sensitive} in the sense  that
	\begin{align}\label{eq:aha}
	p_{i}-p_{i-1}\geq \frac{1}{m}.
	\end{align}
	Indeed, consider $x_j$ such that $h_{x_j}=\arg\min_k \Loss_S(h_{x_k})$,	and let $i = \pos(x_j,S)$. Then,
	\[p_i - p_{i-1} = \frac{\Loss_S(h_{x_j})}{\sum_{i'\leq m}\Loss_S(h_{x_{i'}})} \geq \frac{1}{m}.\]

In the general case we show that any homogeneous algorithm that learns $\H_n$ 
	satisfies \Cref{eq:aha} for \emph{typical} samples (see \Cref{lem:hplAlg}).
	The intuition is that any algorithm that learns the distribution $D$ must output a Gibbs-classifier $Q_S$
	such that for typical points $x$, if $x > n/2$ then $\Pr_{h\sim Q_S}[h(x)=1] \approx 1$,
	and if $x \leq n/2$ then $\Pr_{h\sim Q_S}[h(x)=1]\approx 0$.
	Thus, when traversing all $x$'s from $1$ up to $n$ there must be a jump between $p_{i-1}$ and $p_{i}$
	for some $i$.
 
\paragraph{From Sensitive Indices to a Lower Bound on the KL-divergence.} 
How do sensitive indices imply a lower bound on PAC-Bayes?
	This is the most technical part of the proof.
	The crux of it is a connection between sensitivity and the KL-divergence which we discuss next.
	Consider a sensitive index $i$ and let~$x_j$ be the input example such that $\pos(x_j,S)=i$.
	For $\hat{x}\in \X_n$, let $S_{\hat{x}}$ denote the sample obtained by replacing $x_j$ with $\hat{x}$:
	\[S_{\hat{x}}=\langle (x_1,y_1),\ldots, (x_{j-1},y_{j-1}), (\hat{x}_j,y_j),(x_{j+1},y_{j+1})\ldots (x_m,y_m).\rangle,\]
	and let $Q_{\hat{x}} := Q_{S_{\hat{x}}}$ denote the posterior outputted by $\alg$ given the sample $S_{\hat{x}}$. 
	Consider the set $I\subseteq \X_n$ of all points $\hat{x}$ such that $S_{\hat{x}}$ is equivalent to $S$.
	\Cref{eq:aha} implies that that for every $x,\hat{x}\in I$, 
\begin{align*}
\Pr_{h\sim Q_{\hat{x}}}[h(x)=1] =
\begin{cases}
 p_{i-1} &x<\hat{x} ,\\
p_i	    &x>\hat{x}.
\end{cases}
\end{align*}
	
Combined with the fact that $p_i- p_{i-1}\geq 1/m$, this implies a lower bound on KL-divergence
between an arbitrary prior $P$ and $Q_{\hat{x}}$ for most $\hat x\in I$.
This is summarized in the following lemma:

\begin{lemma}[Sensitivity Lemma]\label{lem:sensitivity}
Let $I$ be a linearly ordered set and let~$\{Q_{\hat{x}}\}_{\hat{x}\in I} $ be a family of posteriors supported on $\{\pm 1\}^I$.
Suppose there are $q_1<q_2\in [0,1]$ such that for every $x,\hat{x}\in I$: 
\begin{align*}
x<\hat{x} &\implies \Pr_{h\sim Q_{\hat{x}}}[h(x)=1]\leq  q_1+ \frac{q_2-q_1}{4},\\
x>\hat{x} &\implies \Pr_{h\sim Q_{\hat{x}}}[h(x)=1]\geq q_2-\frac{q_2-q_1}{4}.
\end{align*}
Then, for every prior distribution $P$, if $\hat x\in I$ is drawn uniformly at random, 
	then the following event occurs with probability at least $1/4$:
	\[\kl{Q_{\hat{x}}}{P}=\Omega\Bigl((q_2-q_1)^2\frac{\log \lvert I\rvert}{\log \log \lvert I\rvert}\Bigr).\]

\end{lemma}
The sensitivity lemma tells us that in the above situation, the KL divergence between $Q_{\hat{x}}$ and any prior $P$, for a random choice $\hat{x}$, scales in terms of two quantities: the distance between the two values, $q_2-q_1$, and the size of $I$. 

The proof of \Cref{lem:sensitivity} is provided in \Cref{prf:sensitivity}.
In a nutshell, the strategy is to bound from below $\kl{Q_{\hat{x}}^r}{P^r}$, where $r$ is sufficiently small;
	the desired lower bound then follows from the chain rule,
	$\kl{Q_{\hat{x}}}{P} = \frac{1}{r}\kl{Q_{\hat{x}}^r}{P^r}$.
	Obtaining the lower bound with respect to the $r$-fold products is the crux of the proof.
	In short, we will exhibit events $E_{\hat{x}}$ such that $Q_{\hat{x}}^r(E_{\hat{x}})\geq \frac{1}{2}$ for every $\hat{x}\in I$, 
	but $P^r(E_{\hat{x}})$ is tiny for $\frac{\lvert I\rvert}{4}$ of the $\hat{x}$'s. 
	This implies a lower bound on $\kl{Q_{\hat{x}}^r}{P^r}$ since 
	\[\kl{Q_{\hat{x}}^r}{P^r}\geq  \kl{Q_{\hat{x}}^r(E_{\hat{x}})}{P^r(E_{\hat{x}})},\]
	by the data-processing inequality.

%\Cref{lem:sensitivity} links the KL-divergence to sensitive indices. 
%This is the technical tool which will be used to derive  the lower bound on the PAC-Bayes quantity.

\paragraph{Wrapping Up.}
We now continue in deriving a lower bound for $\alg$.
Consider an input sample $S\sim D^m$. 
In order to apply \Cref{lem:sensitivity}, fix any equivalence-type $(\pi,y)$ with a sensitive index $i$ and let $x_j$ be such that $\pos(x_j;S)=i$.
The key step is to condition the random sample $S$ on $(\pi,y)$ as well as on $\{x_t\}_{t=1}^m\setminus\{x_j\}$ -- all sample points besides the sensitive point $x_j$.
Thus, only $x_j$ is remained to be drawn in order to fully specify $S$.
Note then, that by symmetry $\hat{x}$ is uniformly distributed in a set $I\subseteq \X_n$, 
and plugging $q_1:=p_i, q_2:=p_{i-1}$ in \Cref{lem:sensitivity} yields that for any prior distribution $P$:
\[\kl{Q_S}{P}\ge \tilde\Omega\left(\frac{1}{{m}^2}\log (\lvert I\rvert)\right) ,\]
with probability at least $1/4$.
Note that we are not quite done since the size $\lvert I\rvert$ is a random variable which depends on the type $(\pi,\bar y)$ and the sample points $\{x_k\}_{k\neq j}$.
However, the distribution of $\lvert I\rvert$ can be analyzed by elementary tools.
In particular, we show that $\lvert I\rvert \geq \Omega(\lvert \X_n\rvert/m^2)$ with high enough probability,
which yields the desired lower bound on the PAC-Bayes quantity. (In the general case $\lvert \X_n\rvert$ is replaced
by the size of the homogeneous set.)
%We omit the analysis of $\lvert I\rvert$ from this exposition and refer the reader to the full version for these details.

\section{Proofs}\label{sec:prfs}
\subsection{Proof of \Cref{thm:main}}

%The following technical Lemma roughly state that for any algorithm $A$ there is a large homoegnous set. 
%This will allow us to reduce the general case to the case of homogeneous algorithms:
%\begin{lemma}\label{lem:homo}
%Let $A$ be an algorithm that is defined over input samples of size $m$ over the domain $\X_n$ and outputs posterior distribution $Q_{S}$. 
%Then? there is a set $\X'\subseteq \X_n$ that is $(\gamma,m)$-homogeneous with respect to $A$ of size 
%\begin{align}\label{eq:size} |\X'|\ge \growth(\gamma,m,n),\end{align}
%\end{lemma}
%The Lemma is a slight strengthening of Lemma 9 in \cite{alon}. The proof is almost identical, for completeness we repeat it in \cref{prf:homo}.
%
%
%\begin{lemma}[Sensitivity Lemma]\label{lem:sensitivity}
%Let $I\subseteq \X$ such that $\lvert I\rvert >16$, and let~$\{Q_{\hat{x}}\}_{\hat{x}\in I} $ be a family of posteriors supported on $\{\pm 1\}^I$.
%Suppose there are $q_1<q_2\in [0,1]$ such that for every $x,\hat{x}\in I$: 
%\begin{align*}
%x<\hat{x} &\implies \Pr_{h\sim Q_{\hat{x}}}[h(x)=1]\leq  q_1+ \frac{q_2-q_1}{4},\\
%x>\hat{x} &\implies \Pr_{h\sim Q_{\hat{x}}}[h(x)=1]\geq q_2-\frac{q_2-q_1}{4}.
%\end{align*}
%Then, for every prior distribution $P$, if $\hat x\in I$ is drawn uniformly at random, 
%	then the following event occurs with probability at least $1/8$:
%	\[\kl{Q_{\hat x}}{P}\ge \sqrt{|q_1-q_2|}\frac{\log |I|}{16\log \log |I|}.\]
%
%\end{lemma}
%
%
%\newpage

Let $A$ be an algorithm as in the premise of \Cref{thm:main}.
That is, $A$ receives as input a labeled sample $S$ of length $m$ and outputs a posterior $Q_{S}$. 
By \Cref{lem:homo}, there exists $\X'\subseteq \X_n$ of size $\lvert \X'\rvert = k \ge \growth(m,\gamma,n)$
such that the restriction of $A$ to inputs from $\X'$ is $\gamma$-approximate $m$-homogeneous.
Without loss of generality, assume that $\X' = \X_k$ consists of the first $k$ points in~$\X_n$
and that $k$ is an even number.

By the definition of approximate homogeneity (\Cref{def:approxhomo}) 
it follows that for every equivalence type $(\pi, \bar y)$, where $\pi$ is a permutation, 
there is a list $(p^{(\pi,\bar y)}_i)_{i=0}^{m}\in [0,1]^{m+1}$ such that for every sample $S\in  (\X_k\times\{0,1\})^m$ whose type 
is $(\pi, \bar y)$ and and every $x\in \X_k\setminus S$:
\[ \bigl\lvert Q_S(x)-p_i^{(\pi, \bar y)}\bigr\rvert \le \frac{\frac{\gamma}{5m}}{2} =  \frac{\gamma}{10m},\]
where $\pos(x,S)=i$.
For the rest of the proof fix $D$ to be the distribution over examples $(x,y)$
such that $x$ is drawn uniformly from $\X_k$ and $y=-1$ if and only if $x\leq k/2$.
%Thus, in particular $\Pr_{(x,y)\sim D}[y=1]=1/2$.
The underlying property we will require is summarized in the following claim:
\begin{claim}\label{lem:hplAlg}
%Let $D$ be a distribution over examples $(x,y)$ such that $x$ is drawn uniformly from $\X_k$ and $\Pr_{(x,y)\sim D}[y=1]=1/2$.
Let $(\pi, \bar y)$ be an equivalence-type, where $\pi$ is a permutation.
Then, one of the following holds: either there exists a sensitive index $0\le i\le m$ such that
\begin{equation}\label{eq:sensitive} |p_{i}^{(\pi, \bar y)}-p_{i-1}^{(\pi, \bar y)}|\ge \frac{\gamma}{2m},\end{equation}
or else, 
\[\Loss_D(Q_S)>\frac{1}{2}-\gamma-\frac{m}{k}\]
with probability $1$ over  $S\sim D^m(\cdot|(\pi,\bar y))$.
\end{claim}
The proof of \Cref{lem:hplAlg} is deterred to \Cref{prf:hplAlg}. 

With \Cref{lem:hplAlg} in hand we proceed with the proof of \Cref{thm:main}. 
Let $S$ be a sample and let $(\pi, \bar y)$ denote its equivalence-type. 
Define an interval $I(S)\subseteq \X_k$ as follows. 
\begin{itemize}
    \item if $\pi$ is not a permutation then $I(S)=\emptyset$. 
    \item If $(\pi, \bar y)$ does not have a sensitive index that satisfies \Cref{eq:sensitive} then $I(S)=\emptyset$.
    \item Finally, if $\pi$ is a permutation and $(\pi, \bar y)$ has a sensitive index $i$ then set\footnote{For concreteness, let $i$ be the minimal sensitive index.}
    \begin{align*}
I(S)= 
\begin{cases}
(x_j^-,x_j^+) & \frac{k}{2}\notin (x_j^-,x_j^+), \\
(x_{j}^-,\frac{k}{2}] & \frac{k}{2}\in (x_j^-,x_j^+) \text{ and } y_j=-1,\\
(\frac{k}{2}, x_{j}^+) & \frac{k}{2}\in (x_j^-,x_j^+) \text{ and } y_j=+1,
\end{cases}
\end{align*}
where $x_j$ is such that $\pos(x_j;S)=i$, and $x_j^-=\max(\{x_t : x_t < x_j \}\cup\{0\})$ and $x_j^+=\min(\{x_t: x_t > x_j\}\cup\{k+1\})$.

\end{itemize}

We next define two events which will be used to finish the proof. 
First, consider the event that the drawn sample $S$ satisfies either\footnote{We use here 
	the convention, that $\frac{\log x}{\log \log x}=-\infty$ for $x\leq 2$. 
	Alternatively, one can assume that \Cref{eq:kl} holds vacuously if $|I(S)|=0$}
\begin{equation}\label{eq:kl} 
\kl{Q_S}{P} = 
\Omega\Bigl(\frac{\gamma^2}{m^2}\frac{\log \lvert I(S)\rvert }{\log\log \lvert I(S)\rvert }\Bigr),
\end{equation}
 or
\begin{equation}\label{eq:loss}
\Loss (Q_{S})\ge \frac{1}{2}-\gamma-\frac{m}{k},
\end{equation}
We show that this event occurs with probability at least $1/4$:
\begin{claim}\label{cl:2}
Define $E_1$ to be the event
\[E_1= \Bigl\{S\in (\X_k\times\{\pm 1\})^m : S\text{ satisfies \Cref{eq:kl} or \Cref{eq:loss}}\Bigr\}.\] 
Then, $E_1$ occurs with probability at least $1/4$ over $S\sim D^m$.
\end{claim}
The proof of \Cref{cl:2} is deterred to \Cref{prf:cl2}. 
The second event we consider is that the drawn sample $S$ satisfies either \Cref{eq:loss} or
\begin{align}\label{eq:last}
\lvert I(S)\rvert \geq \frac{\growth(m,\gamma,n)}{8(m+1)^2}.
\end{align}
We show that this event occurs with probability at least $7/8$:
\begin{claim}\label{cl:3}
Define $E_2$ to be the event
\[ E_2= \{S: \textrm{S~satisfies \Cref{eq:loss}~or~\Cref{eq:last}}\}.\] 
Then $E_2$ occurs with probability at least $7/8$ over $S\sim D^m$
\end{claim}
The proof of \Cref{cl:3} is deterred to \Cref{prf:cl3}. 
With \Cref{cl:2,cl:3} in hand, the proof of \Cref{thm:main} is completed as follows. 
First, a union bound implies that the event $E_1\cap E_2$ occurs with probability at least $1/16$. 
That is, with probability at least $1/16$ either \Cref{eq:loss} holds and we are done, 
or else, if \Cref{eq:loss} doesn't hold, then both \Cref{eq:kl,eq:last} hold simultaneously,
which yields that
\begin{align*}\kl{Q_S}{P}&\ge
\Omega\Bigl(\frac{\gamma^2}{m^2}\frac{\log \lvert I(S)\rvert }{\log\log \lvert I(S)\rvert }\Bigr) 
\tag{By \Cref{eq:kl}}
\\
&\ge \Omega\Bigl(\frac{\gamma^2}{m^2}\frac{\log \frac{\growth(m,\gamma,n)}{8(m+1)^2}}{\log\log \frac{\growth(m,\gamma,n)}{8(m+1)^2}}\Bigr). 
\tag{By \Cref{eq:last}}
\end{align*}
This concludes the proof of \Cref{thm:main}. 

\qed

We are thus left with proving \Cref{lem:hplAlg,cl:2,cl:3}.

\subsubsection{Proof of \Cref{lem:hplAlg}}\label{prf:hplAlg}
%Let $A$ be an $\gamma$-homogeneous posterior-learning algorithm over the domain $\X_k$. 
%Fix an order type $(\pi,\bar y)$ and consider $S\sim D^m(\cdot|(\pi, \bar y)$. 
%Note that since $k$ is even, it follows that $\Pr_{(x,y)\sim D}[y=1]=1/2$. 
Let $(\pi, \bar y)$ be an equivalence-type such that $\pi$ is a permutation.
Assume that 
\begin{align}\label{eq:smallloss}
\Loss(Q_S)<\frac{1}{2}-\gamma-\frac{m}{k} \
\end{align}
occurs with a positive probability over $S\sim D^m(\cdot \vert \pi,\bar y)$.
We first show that there is~$i$ such that 
\begin{align}\label{eq:i1i2}|p^{\pi,\bar y}_{i}-p^{\pi,\bar y}_{0}|> \gamma/2.\end{align} 
Indeed, assume the contrary and fix a sample $S$ with type $(\pi, \bar y)$ which satisfies \Cref{eq:smallloss}.
Recall that $A$ is homogeneous, hence for every $x\notin \underline{S}$, 
\[ |Q_S(x)-p_i^{\pi, \bar y}|<\frac{\gamma}{10m},\]
where $i=\pos(x,S)$.
On the other hand, since \Cref{eq:i1i2} is not met by any~$i$, it follows that for every $x\notin \underline{S}$:
\begin{align*}|Q_{S}(x) -p_0^{\pi, \bar y}|&= |Q_{S}(x)-p_i^{\pi, \bar y}+p_i^{\pi, \bar y} -p_0^{\pi, \bar y}|\\
&\le  |Q_{S}(x)-p_i^{\pi, \bar y}|+|p_i^{\pi, \bar y} -p_0^{\pi, \bar y}|\\
&\le   \frac{\gamma}{10m}+\frac{\gamma}{2}\\
&\le \gamma.
\end{align*}
Thus, $\Pr_{h\sim Q_S}[h(x)=1]\in [p_0^{\pi,\bar y}-\gamma, p_0^{\pi,\bar y}+\gamma ]$, for every $x\in \X_{k}\setminus\underline{S}$.
Now, since $\Pr_{(x,y)\sim D}[y=1]=1/2$ it follows that
\begin{align*} \Loss_D(Q_S)&\ge \frac{1}{2}-\gamma-\frac{m}{k}.
\end{align*}
Indeed, for every $x\notin S$, if $x\leq k/2$ then $h\sim Q_S$ errs on $x$ with probability
at least $q_1=p_0^{\pi,\bar y}-\gamma$, and if $x> k/2$
then $h\sim Q_S$ errs on $x$ with probability at least $q_2=1-(p_0^{\pi,\bar y}+\gamma)$.
Thus, the expected loss of $h\sim Q_S$ conditioned on $x\notin \underline{S}$ is at least $\frac{q_1+q_2}{2}=1/2-\gamma$,
and the above inequality follows by taking into account that $h\sim Q_S$ may have zero error on the $m$ points in $S$.

Finally, let $i$ be some index that satisfy \Cref{eq:i1i2}, then because $0\le i\le m$ we obtain via telescoping that there must be some $i'\leq i$, such that 
\[ |p_{i'}^{\pi, \bar y}- p_{i'-1}^{\pi, \bar y}|\ge \frac{\gamma}{2m}.\]
\qed

\subsubsection{Proof of \Cref{cl:2}}\label{prf:cl2}
\begin{proof}[\textbf{Proof of \Cref{cl:2}}]

It is enough to show that $E_1$ occurs with probability at least $1/4$ over $S\sim D^m(\cdot|\pi,\bar y)$ 
for every fixed equivalence-type $(\pi, \bar y)$.
Indeed, by summing over all equivalence types, the law of total probability then implies that $E_1$ occurs with probability at least $1/4$ over $S\sim D^m$.

Fix an equivalence-type $(\pi,\bar y)$. We may assume that $\pi$ is a permutation and that $(\pi, \bar y)$
    has a sensitive index $i$ (or else \Cref{eq:kl} trivially holds by the definition of $I(S)$ and we are done).
    If \Cref{eq:loss} holds with probability at least $1/4$ then also $E$ occurs with probability at least $1/4$ and we are done.
    Thus, assume that \Cref{eq:loss} holds with probability less than $1/4$.
    It suffices to show that \Cref{eq:kl} holds with probability at least $1/4$.
	By \Cref{lem:hplAlg}, there is a sensitive index $i$ such that
\begin{align*}
\lvert p_{i}^{(\pi,\bar y)}-p_{i-1}^{(\pi,\bar y)} \rvert> \frac{\gamma}{2m}.
\end{align*}
Let $x_j$ in $S$ be such that $\pos(x_j;S)=i$.
It will be convenient to consider the following (slightly convoluted) process of sampling a pair of (correlated) samples from $D^{m}(\cdot|\pi,\bar y)$:
\begin{enumerate}
\item Sample $T=\langle(x_1,y_1)\ldots (x_m,y_m)\rangle\sim D^m(\cdot \vert \pi, \bar y)$.
\item Resample only the sensitive point $x_j$ while keeping all other points fixed, as well as the equivalence type $(\pi, \bar y)$. Let $\hat x$ denote the newly sampled point and let $T_{\hat{x}}$ denote the sample obtained
by replacing $x_j$ by $\hat{x}$.
\item Set $S=T_{\hat{x}}$
\end{enumerate}
Note that both $T$ and $S$ are drawn from $D^m(\cdot \vert \pi, \bar y)$ and that $I(T)=I(S)$ always.
Since the marginal distribution of $D$ is uniform over $\X_k$, 
by symmetry it follows that the point $\hat{x}$ drawn in Step 2 is uniform in the interval $I(T)=I(S)$. 
%For $\hat{x}\in I(S')$, let $S_{\hat{x}}$ be the sample obtained when $x_j= \hat{x}$.
%Note that sampling $S\sim D^m(\cdot \vert \pi, \bar y)$ conditioned on $\{x_t\}_{t=1}^m\setminus\{x_j\}$
%is equivalent to sampling $S_{\hat{x}}$, where $\hat{x}$ is  drawn uniformly from $I$.
Our next step is to apply \Cref{lem:sensitivity} on the family of distributions $\{Q_{T_{\hat{x}}}\}_{\hat{x}\in I(T)}$.
Towards this end, we first fix $T$  and show that the premise of \Cref{lem:sensitivity} is satisfied, with $I=I(T)$,  $q_1=p_{i-1}^{(\pi,\bar y)}$ and $q_2=p_i^{(\pi, \bar y)}$.\footnote{Here we assume without loss of generality that $p_{i-1}^{(\pi,\bar y)}< p_i^{(\pi, \bar y)}$. If the reverse inequality holds then the argument follows by applying \Cref{lem:sensitivity} with respect to the reverse linear order over $I(T)$.}
Indeed, by homogeneity it follows that for each $x\in I(T)$, if $x<\hat{x}$ 
\begin{align*}
\Bigl\lvert \Pr_{h\sim Q_{\hat{x}}}[h(x)=1]-p^{\pi,\bar y}_{i-1}\Bigr\rvert &\le \frac{\gamma}{10m}\\
					&< \frac{\lvert p_{i}^{(\pi,\bar y)}-p_{i-1}^{(\pi,\bar y)}\rvert}{4}, \tag{because $i$ is sensitive}
\end{align*}
and similarly if $x\ge \hat{x}$:
\[\Bigl\lvert\Pr_{h\sim Q_{\hat{x}}}[h(x)=1] -p^{(\pi,\bar y)}_{i}\Bigr\rvert < \frac{\lvert p_{i}^{(\pi,\bar y)}-p_{i-1}^{(\pi,\bar y)}\rvert}{4}.\]
Thus, applying \Cref{lem:sensitivity} on the family of distributions $\{Q_{T_{\hat{x}}}\}_{\hat{x}\in I(T)}$ yields that for every $T$ sampled in Step 1,
the following holds with probability at least $1/4$ over sampling $\hat{x}$:
\begin{align*}
\kl{Q_{S}}{P}&=
\kl{Q_{T_{\hat{x}}}}{P}
\\
&\ge 
\Omega\Big(\bigl( p^{(\pi,\bar y)}_{i-1}-p^{(\pi,\bar y)}_{i}\bigr)^2\frac{\log (\lvert I(T)\rvert)}{\log\log \lvert I(T)\rvert}\Bigr)
\\
&\ge \Omega\Bigl(\frac{\gamma^2}{m^2}\frac{\log \lvert I(T)\rvert}{\log\log \lvert I(T)\rvert}\Bigr)
\\
& = 
\Omega\Bigl(\frac{\gamma^2}{m^2}\frac{\log \lvert I(S)\rvert}{\log\log \lvert I(S)\rvert}\Bigr).
\end{align*}
Note that the above holds for any fixed $T$. 
Taking expectation over $T$ it follows that with probability at least $1/4$ over $S\sim D(\cdot|(\pi, \bar y))$,
\[\kl{Q_S}{P}\ge 
\Omega\Bigl(\frac{\gamma^2}{m^2}\frac{\log \lvert I(S)\rvert}{\log\log \lvert I(S)\rvert}\Bigr).\]
As discussed, taking expectation over the equivalence type concludes the proof.
\end{proof}

\subsubsection{Proof of \Cref{cl:3}}\label{prf:cl3}
\begin{proof}[\textbf{Proof of \Cref{cl:3}}]

Consider $S\sim D^m$ where $S=\langle(x_1,y_1),\ldots (x_m,y_m) \rangle$.
We claim that with probability at least $7/8$, every two unlabeled examples $x_i,x_j$ with $i\neq j$ are at distance at least $\frac{k}{8(m+1)^2}$ from each other and from $k/2$. 
Indeed, fix any distinct $x',x''\in \{x_1,\ldots, x_m,k/2\}$. 
Recall that the distribution~$D$ satisfies that $x_1,\ldots,x_m$ are sampled uniformly and ind.\ from $\X_k$.
Thus, the probability that $0\le x'-x''<\frac{k}{8 (m+1)^2}$ is at most $\frac{1}{8(m+1)^2}$. 
A union bound over all possible ${m+1 \choose 2}$ pairs implies that 
that the following holds with probability at least $\frac{7}{8}$ over $S\sim D^m$:
\begin{equation}\label{eq:large}
\Bigl(\forall\text{ distinct } x',x''\in \Bigl\{x_1,\ldots, x_m,\frac{k}{2}\Bigr\}\Big): \lvert x'-x''\rvert\geq \frac{k}{8(m+1)^2}.
\end{equation}
We will now show that the latter event implies $E_2$.
Let $S$ be a sample satisfying \Cref{eq:large}.
In particular, $x_i\neq x_j$ for every distinct $i,j\leq m$ and so the order-type $\pi=\pi(S)$ is a permutation. 
Now, if $S$ satisfies \Cref{eq:loss} then $S\in E_2$ and we are done.
Else, by \Cref{lem:hplAlg} there exists a sensitive index that satisfies \Cref{eq:sensitive}
and therefore $I(S)=(x',x'')$, where $x',x''$ are distinct points in $\{x_1\ldots,x_m, k/2\}$. 
Thus, \[|I(S)|\ge \frac{k}{8(m+1)^2},\]
and \Cref{eq:last} holds, which also gives $S\in E_2$.
Thus, every $S$ which satisfies \Cref{eq:large} is in $E_2$ 
and so $E_2$ occurs with probability at least $7/8$. 
%
%To summarize, with probability $7/8$, there are no repetitions in $S$ and either \cref{eq:loss} holds or there exists a sensitive index and by definition: \begin{align}\label{eq:is}|I(S)|\ge \frac{k}{8m^2}\ge \frac{\growth(\gamma,m,n)}{8m^2}\end{align}
\end{proof}

\subsection{Proof of \Cref{lem:homo}}\label{prf:homo}

We next prove \Cref{lem:homo} which  
establishes the existence of a ``largish'' homogeneous set with respect to  an arbitrary algorithm $A$.

\paragraph{Notation.}
%Let $S=\langle(x_i,y_i)\rangle_{i=1}^m$ be a sample of size $m$ and let $(\pi, \bar y)$ denote the equivalence-type of $S$.
%Note that $\lvert \{x_i: i\leq m\}\rvert$ -  the number of distinct points in $S$ equals to $\max_i \pi(i)$.
%In particular equivalent samples have the same number of distinct points.
%Thus,  we define the {\it effective-size} of an equivalence-type $(\pi,y)$ to be $\max_i \pi(i)$ and denote it by $\sigma(\pi)$.
%For $k=1,\ldots,m$, let $\mathcal{T}_k=\{(\pi, \bar y): \sigma(\pi)=k\}$ be the class of all types whose effective-size is $k$.
%
Recall from \Cref{eq:pos} the definition of $\pos(x,S)$ which was defined for a sample $S$ and a point $x$.
It will be convenient to extend this definition to sets:
for $R\subseteq \X_n$ and $x\in \X_n$ define $\pos(x,R)= \lvert\{x'\in R : x'\leq x \}\rvert$.

\paragraph{From Sets to Samples.}
Let $(\pi, \bar y)$ be an equivalence-type whose order-type is a permutation and let
 $D=\{x_1<\ldots <x_m\}\subseteq \X_n$ be a set of $m$ points.
Denote by $D^{\pi, \bar y}=\langle(x_{i_j}, y_{i_j})\rangle_{j=1}^m$ the sample 
obtained by ordering and labeling the elements of $D$ such that $D^{\pi, \bar y}$ has type $(\pi, \bar y)$;
that is, $D^{\pi, \bar y}$ is defined such that for every $j\leq m$,
\begin{equation}\label{eq:setsample}
\pi(j)=\pos(x_{i_j},D^{\pi, \bar y}) =\pos(x_{i_j},D) ~\text{ and }~ \bar y =(y_1,\ldots, y_m).
\end{equation}
%
%define the algorithm $A^{\pi, \bar y}$ which takes \underline{subsets} $\bar S\subseteq X$ 
%of size $m'=\rho(s)$ to Gibbs-Classifiers $Q_{\bar S}^{\pi, \bar y}$ as follows:
%given $\bar S=\{x_1 <\ldots < x_{m'}\}\subseteq \X$, 
%The Gibbs-classifier $Q_{\bar S}^{\pi, \bar y}$ is defined to be the output of $A$ on the input sample $S^{\pi, \bar y}$.

\paragraph{A Coloring.}
We define a coloring over subsets $D\subseteq \X_n$ of size $\rvert D\rvert = m+1$.
Let $D=\{x_0<x_1<\ldots<x_{m}\}$ be a $(m+1)$-subset of~$\X_n$.
The coloring assigned to $D$ is 
\[ c(D) = \bigl\{(p_0^{\pi, \bar y},\ldots,p_{m}^{\pi, \bar y}) : (\pi, \bar y) \text{ is an equivalence-type s.t.\ $\pi$ is a permutation}\bigr\},\]
where each $p_i^{\pi, \bar y}$ is defined as follows:
let $D_{-i} = D\setminus\{x_i\}$. 
For each equivalence type $(\pi, \bar y)$ such that $\pi$ is a permutation
consider the sample $D_{-i}^{\pi, \bar y}$ (see \Cref{eq:setsample}),
and define $p_i^{\pi, \bar y}$ to be the fraction of the form~$\frac{t\cdot\gamma}{10m}$ for $t\in\N$ which is closest to 
\[\Pr_{h\sim Q_{-i}^{\pi, \bar y}}[h(x_i)=1],\]
where $Q_{-i}^{\pi, \bar y}$ is the stochastic classifier obtained by applying $A$ on $D_{-i}^{\pi, \bar y}$.

Since the total number of equivalence-types whose order-type is a permutation is at most $m!\cdot 2^m$, it follows that
the total number of colors is at most~$m!\cdot 2^m\cdot \lceil\frac{10m}{\gamma}+1\rceil^{(m+1)} \leq (\frac{100m}{\gamma})^{2m}$.

\paragraph{Ramsey.}
We next apply Ramsey Theorem to derive a large $\X'\subseteq \X_n$ such that every subset $D\subseteq \X_n$ of size $m+1$ has the same color.
Later we will argue that $A$ is $\gamma$-approximately homogeneous with respect to $\X'$ which will finish the proof.

We will use the following quantitative version of Ramsey Theorem due to~\cite{erdos52combinatorial} 
(see also the book \citep{graham90ramsey}, or Theorem 10.1 in the survey by~\cite{mubayi17survey}).
Here, the {\it tower function} $\twr_k(x)$ is defined by the recursion
\[
\twr^{(i)} x = 
\begin{cases}
x &i = 1,\\
2^{\twr{(i-1)}(x)} &i> 1. 
\end{cases}
\]
\begin{theorem}[Ramsey Theorem \citep{erdos52combinatorial}]\label{thm:ramsey}
Let $s>t\geq 2$ and $q$ be integers, and let 
\[N\geq \twr_t(3sq\log q).\] 
Then, for every coloring of the subsets of size $t$ of a universe of size $N$ using $q$ colors
there is a homogeneous subset\footnote{A subset of the universe is homogeneous if all of its $t$-subsets have the same color.} of size $s$.
\end{theorem}
Stated differently, \Cref{thm:ramsey} guarantees the existence of a homogeneous subset of size
\begin{equation}\label{eq:ramsey}
\frac{\log^{(t-1)}(N)}{3q\log q}.
\end{equation}
%We next apply \Cref{thm:ramsey} iteratively to obtain a chain 
%\[\X=\X_0\supseteq \X_1\supseteq \X_2\supseteq\ldots \supseteq \X_m=\X'\]
%such that for every $i\leq m$, $\X_i$ is homogenous with respect to all subsets $D\subseteq \X$ of size at most $i+1$. have the same color.
%Indeed, $\X_{i}$ is obtained by applying \Cref{thm:ramsey} on $\X_{i-1}$ with respect to the coloring of all $i+1$ subsets. 
Thus, by plugging $q:=(\frac{10m}{\gamma})^{2m},t:=m+1,N:=n$ in \Cref{eq:ramsey} we get
a homogeneous set $\X'\subseteq \X_n$ of size
\[\lvert \X'\rvert \geq \frac{\log^{(m)}(n)}{3(\frac{10m}{\gamma})^{2m} \cdot 2m\log (\frac{10m}{\gamma})}
\geq \frac{\log^{(m)}(n)}{(\frac{10m}{\gamma})^{3m} }.\]
%\[\lvert\X_{i} \rvert \geq  \frac{\log^{(i)}(\lvert \X_{i-1}\rvert)}{3(\frac{10m}{\gamma})^{2m} \cdot 2m\log (\frac{10m}{\gamma})}
%\geq \frac{\log^{(i)}(\lvert \X_{i-1}\rvert)}{(\frac{10m}{\gamma})^{3m} }.\]
%By induction, this implies that $\X'=\X_m$ satisfies
%\[
%\lvert\X'\rvert \geq  \frac{\log^{(m^2)}(n)}{(\frac{10m}{\gamma})^{3m^2}}.
%\]

\paragraph{Wrapping-up.}
It remains to show that $A$ is $\gamma$-approximately homogeneous with respect to $\X'$.
By the construction of $\X'$ there exist a specific color 
\[L=\{(p^{\pi, \bar y}_i)_{i=0}^{m} : (\pi, \bar y) \text{ is an equivalence-type s.t.\ $\pi$ is a permutation}\}\]
such that $c(D)=L$ for every $D=\{x_0<\ldots <x_m\}\subseteq \X'$.
We need to show that for every pair of equivalent samples $S',S''$ whose order-type is a permutation and for 
every $x\in \X'\setminus\underline{S}, x'\in \X'\setminus\underline{S'}$ such that $\pos(x,S)=\pos(x',S')$:
\[ \Bigl\lvert \Pr_{h\sim Q_S}[h(x)=1] - \Pr_{h'\sim Q_{S'}}[h'(x')=1]\Bigr\rvert \leq \frac{\gamma}{5m}.\]
Let $(\pi, \bar y)$ be an equivalence-type such that $\pi$ is a permutation,
let $S$ be any sample whose equivalence-type is $(\pi, \bar y)$, and let $x\in \X'\setminus \bar{S}$.
Consider the set $D=\{x_j: j\leq m\}\cup\{x\}$ and set $i=\pos(x,S)$.
By the definition of $D_{-i}^{\pi, \bar y}$, we have $D_{-i}^{\pi, \bar y}=S$ and hence 
by the definition of $p^{\pi, \bar y}_i$  we have
\[\Bigl\lvert \Pr_{h\sim Q_S}[h(x)=1] - p^{\pi, \bar y}_i\Bigr\rvert=\Bigl\lvert \Pr_{h\sim Q^{\pi, \bar y}_{-i}}[h(x)=1] - p^{\pi, \bar y}_i\Bigr\rvert\leq \frac{\gamma}{10m}.\]
Since the latter holds for every sample $S$ whose order type is $(\pi, \bar y)$ and every $x\notin \bar{S}$,
it follows that for every pair of samples $S,S'$ whose order-type is $(\pi,\bar y)$ and every $x\in \X'\setminus\underline{S}, x'\in \X'\setminus\underline{S'}$ such that $\pos(x,S)=\pos(x',S')$:
\begin{align*}
&\Bigl\lvert \Pr_{h\sim Q_S}[h(x)=1] - \Pr_{h'\sim Q_{S'}}[h'(x')=1]\Bigr\rvert \leq \\
&\Bigl\lvert \Pr_{h\sim Q_S}[h(x)=1] - p^{\pi, \bar y}_i\Bigr\rvert + \Bigl\lvert \Pr_{h\sim Q_S}[h(x)=1] - p^{\pi, \bar y}_i\Bigr\rvert \leq \frac{\gamma}{10m} + \frac{\gamma}{10m}=\frac{\gamma}{5m},
\end{align*}
where $i:=\pos(x,S)=\pos(x',S')$. This finishes the proof.
\qed

\subsection{Proof of \Cref{lem:sensitivity}}\label{prf:sensitivity}
\paragraph{Notation.}
We will assume without loss of generality that $I=\{ 1,2,3,...,\lvert I\rvert\}$.
	Also, to simplify the presentation, we will assume that $\lvert I\rvert$ is a power of $2$, 
	i.e.\ $\lvert I\rvert=2^b$ for some $b\in\mathbb{N}$.
	(Removing this assumption is straight-forward, but complicates some of the notation.)

\paragraph{Overview.}
Let $P$ be an arbitrary prior supported on $\{\pm 1\}^I$. 
	Our goal is to show that at least~$\lvert I\rvert/4$ of all $\hat{x}$'s in $I$ satisfy
	\[\kl{Q_{\hat x}}{P}\ge \Omega\Bigl((q_2-q_1)^2\frac{\log \lvert I\rvert}{\log \log \lvert I\rvert}\Bigr) = \Omega\Bigl((q_2-q_1)^2\frac{b}{\log (b)}\Bigr).\]
	The proof strategy is to bound from below $\kl{Q_{\hat{x}}^m}{P^m}$, where $m$ is sufficiently small;
	the desired lower bound then follows from the chain rule:
	\[\kl{Q_{\hat{x}}}{P} = \frac{1}{m}\kl{Q_{\hat{x}}^m}{P^m}.\]
	Obtaining the lower bound with respect to the $m$-fold products is the crux of the proof.
	In a nutshell, we will exhibit events $E_{\hat{x}}$ such that for every $\hat{x}\in I$, $Q_{\hat{x}}^m(E_{\hat{x}})\geq 1/2$, ,
	but for $\lvert I\rvert/4$ of the $\hat{x}$'s, $P^m(E_{\hat{x}})$ is tiny. 
	This implies a lower bound on $\kl{Q_{\hat{x}}^m}{P^m}$ since 
	\[\kl{Q_{\hat{x}}^m}{P^m}\geq  \kl{Q_{\hat{x}}^m(E_{\hat{x}})}{P^m(E_{\hat{x}})},\]
	by the data-processing inequality.

\paragraph{Construction of The Events $E_{\hat{x}}$.}
For every Gibbs-classifier $Q\in\{Q_{\hat{x}}: \hat{x}\in I\}\cup\{P\}$
	define its \emph{rounded-hypothesis} $\vv_{Q}:X\to \{\pm 1\}$ as follows:
\[\vv_{Q}(x)=
\begin{cases}
-1 & \E_{h\sim Q_{\hat{x}}}[h(x)]\le \frac{q_1+q_2}{2},\\
+1 & \E_{h\sim Q_{\hat{x}}}[h(x)]> \frac{q_1+q_2}{2}.
\end{cases}\]
To simplify notation, let $\vv_{\hat{x}}=\vv_{Q_{\hat{x}}}$.
Note that by the assumption of \Cref{lem:sensitivity}:
\begin{equation}\label{eq:threshold}
\vv_{\hat{x}}(x) =
\begin{cases}
-1 & x< \hat{x},\\
+1 & x>\hat{x}.
\end{cases}
\end{equation}
In words, each $\vv_{\hat{x}}$ is a threshold with a sign-change either right before $\hat{x}$ or right after it.
Next, given $\hh:I\to\{\pm 1\}$, consider the following iterative process which applies binary-search on $\hh$ towards detecting a pair of subsequent coordinates 
which contain a sign-change.

\begin{center}
\noindent\fbox{
\parbox{.99\columnwidth}{
	\begin{center}\underline{\bf Binary-Search}\end{center}
Input: $\vv:I\to\{\pm 1\}$.
\begin{enumerate}
\item Set $I_0=[a_0,b_0]$, where $a_0=0,b_0=\lvert I\rvert=2^b$.
\item For $j=0,\ldots$
\begin{enumerate}
\item If $\lvert I_j\rvert  \leq 2$ then output $I_j$.
\item Query the coordinate $\vv(m_j)$, where $m_j=\frac{a_j+b_j}{2}$.
\item If $\vv(m_j)=+1$ then set $a_{j+1}=a_j,b_{j+1}=m_j$,
\item Else, set $a_{j+1}=m_j+1, b_{j+1}=b_j$.
\end{enumerate}
\end{enumerate}
     }}
\end{center}
The following observations follow from the standard analysis of binary-search.
\begin{enumerate}
\item The process ends after $b-1$ iterations and each of the points $m_j$ queried in Item (b) are even numbers.
\item If the process is applied on a threshold $\vv$ which changes sign from $-$ to $+$ between $x$ and $x+1$
then the output interval $I_{out}$ is $\{x,x+1\}$.
Thus, by \Cref{eq:threshold}, if we apply this process on $\hh=\hh_{\hat{x}}$ then $\hat{x}\in I_{out}$.
\end{enumerate}

Given a sequence of hypotheses $h_1,\ldots,h_m:I\to\{\pm 1\}$, 
	define the \emph{empirical rounded-hypothesis} $\vv_{h_{1:m}}$  by:
\[\vv_{h_{1:m}}(x)=\begin{cases}
-1 & \frac{1}{m}\sum_{i=1}^m \mathbf{1}[h_i(x)=1] \le \frac{q_1+q_2}{2}, \\
+1 &\frac{1}{m}\sum_{i=1}^m \mathbf{1}[h_i(x)=1] > \frac{q_1+q_2}{2}.
\end{cases}
\]
Consider $h_1,\ldots,h_m\sim Q_{\hat{x}}$ for an odd $\hat{x}\in I$.
The following claim shows that with high probability,  
applying the binary search on $\vv_{h_{1:m}}$ yields an output interval $I_{out}$
such that $\hat{x}\in I_{out}$.

\begin{claim}\label{c:binary}
Let $\hat{x}\leq 2^b$ be an odd number. %with $m> \sqrt{\frac{\ln 100b}{q_2-q_1}}$.
Let $J_{out}$ denote the interval outputted by applying the binary search on $\hh_{\hat{x}}$
and let $I_{out}$ denote the interval outputted by applying the binary search on $\hh_{h_{1:m}}$,
where $h_1,\ldots h_m\sim Q_{\hat{x}}$ are drawn independently.
Then, 
\[\Pr_{h_1\ldots h_m\sim Q_{\hat{x}}^m}[I_{out} \neq J_{out}]\leq  b\cdot\exp\Bigl(-\frac{m}{2}(q_2-q_1)^2\Bigr).\]
In particular, if $m = \frac{2(\ln(b)+2)}{(q_2-q_1)^2}$ then $\Pr[\hat{x}\notin I_{out}]\leq \frac{1}{2}$.
\end{claim}
\begin{proof}
Let $x_1,\ldots x_2, \ldots, x_{b-1}$ be the coordinates queried by the binary search on $J_{out}$.
%Observe that since $\hat{x}$ is odd, we have $\hat{x}\neq x_i$ for every $i\leq b-1$.
We will show that with high probability $\hh_{h_{1:m}}(x_i)= \hh_{\hat{x}}(x_i)$ for every $i$,  
which implies that $J_{out}=I_{out}$.
Let $i\leq b-1$ and define 
\[\mu_i=\E_{h\sim Q_{\hat{x}}}[{\bf 1}[h(x_i)=+1]] = \Pr_{h\sim Q_{\hat{x}}}[h(x_i)=+1].\]
Note that $\hat{x}\neq x_i$ (because $x_i$ is even and $\hat{x}$ is odd). 
Therefore, by the assumption of \Cref{lem:sensitivity}: 
\[
\mu_i 
\begin{cases}
\leq \frac{q_2+q_1}{2}-\frac{q_2-q_1}{4}  &x_i < \hat{x},\\
\geq \frac{q_2+q_1}{2}+\frac{q_2-q_1}{4} &x_i > \hat{x}.
\end{cases}
\]
Hence, by a Chernoff bound:
\begin{align*}
\Pr_{h_1\ldots h_m}[\hh_{h_{1:m}}(x_i)  \neq \vv_{\hat{x}}(x_i)] 
&\leq 
\Pr_{h_1\ldots h_m}\Bigl[\frac{1}{m}\sum_{j=1}^m \mathbf{1}[h_j(x_i)=1] \geq \mu_{i} + \frac{q_2-q_1}{4}\Bigr]\\
&\leq \exp\Bigl(-\frac{m}{2}(q_2-q_1)^2\Bigr) \tag{Chernoff Bound}
\end{align*}
Thus, by taking a union bound over all $i\leq b-1$ 
it follows that $\hh_{h_{1:m}}(x) = \hh_{\hat{x}}(x)$ for every $i\leq b-1$
with probability at least $1-\log(\lvert I\rvert)\cdot\exp(-\frac{m}{2}(q_2-q_1)^2)$. 
In particular, with the above probability we have that $J_{out}=I_{out}$.

Lastly, assume $m = \frac{2(\ln(b)+2)}{(q_2-q_1)^2}$. Then, $b\cdot\exp(-\frac{m}{2}(q_2-q_1)^2)\leq 1/2$,
and therefore $\Pr[J_{out}=I_{out}]\geq \frac{1}{2}$. 
Since $\hh_{\hat{x}}$ is a threshold which changes sign either right before $\hat{x}$ or right after~$\hat{x}$, 
it follows that $\hat{x}\in J_{out}$, and therefore $\Pr[\hat{x}\in I_{out}]\geq 1/2$.
\end{proof}

We are now ready to define the events $E_{\hat{x}}$.
Set $m=\frac{2(\ln(b)+2)}{(q_2-q_1)^2}$, according to \Cref{c:binary},
and let $E_{\hat{x}}$ denote the event that $\hat{x}\in I_{out}$.
That is, $E_{\hat{x}}$ is the set of all sequences $h_1,\ldots h_m$
such that $\hat{x}\in I_{out}$, where $I_{out}$ is the 
interval outputted by the binary-search on $\hh_{h_{1:m}}$.
Thus, \Cref{c:binary} says that $Q_{\hat{x}}^m(E_{\hat{x}})\geq \frac{2}{3}$ for an odd $\hat{x}$.

\paragraph{Bounding the KL-divergence.}
We next use the events $E_{\hat{x}}$ to lower bound $\kl{Q_{\hat{x}}}{P}$:
\begin{align*} 
\kl{Q_{\hat{x}}}{P}&= \frac{1}{m}\kl{Q^m_{\hat{x}}}{P^m} \tag{Chain Rule}\\
&\ge  \frac{1}{m}\kl{Q_{\hat{x}}^m(E_{\hat{x}})}{P^m(E_{\hat{x}})} \tag{Data Processing Ineq.}\\
&\ge \frac{1}{m}\left(-\frac{2}{3}\log\Bigl(\frac{2}{3}\Bigr) - \frac{1}{3}\log\Big(\frac{1}{3}\Bigr)-\frac{2}{3}\log\bigl(P^m(E_{\hat{x}})\bigr) \right)\\
\\ &\ge \frac{-\log\bigl(P^m(E_{\hat{x}})\bigr) - 1 }{2m} \label{eq:ineq1}
\end{align*}

Therefore, to lower bound $\kl{Q_{\hat{x}}}{P}$
it suffices to shows that $P^m(E_{\hat{x}})$ is small.
We next establish this for $1/4$ of the $\hat{x}$'s in $I$.
Note that whenever $\hat{x}_1,\hat{x}_2\in I$ are odd and distinct then $E_{\hat{x}_1}\cap E_{\hat{x}_2} = \emptyset$.
Indeed, this follows since the outputted interval $I_{out}$ is of size $\leq 2$ and hence contains at most one odd number.
Thus,
\[\sum_{\hat{x} \text{ is odd}}P^m(E_{\hat{x}})\leq 1.\]
In particular, since there are $2^{b-1}$ odd numbers in $I$,
at least $1/2$ of them must satisfy $P^m(E_{\hat{x}})\leq \frac{1}{2^{b-2}}$.
%
%Next, let $\hat{x}$ is randomly and uniformly chosen from the interval $I$. 
%Then 
%\begin{align*}\E_{\hat{x}\sim U}(P^t(X_{\hat{x}}(\vv)=1))&=
%\frac{1}{|I|}\sum_{\hat{x}\in I} P^t(X_{\hat{x}}(\vv)=1)\\
%&= \frac{1}{|I|}P^t\left(\mathop\bigcup_{\hat{x}\in I}\left\{\vv: X_{\hat{x}}(\vv)=1\right\}\right)\\
%&\le
%\frac{1}{|I|}\\
%&= 
%2^{-b}\end{align*}
%By Markov's inequality then, we obtain that with probability at least $7/8$ over $\hat{x}$:
%\[P^{t}(X_{\hat{x}}(\vv)=1)\le 2^{-b+3}\]
Taken together we obtain that at least $1/4$ of all $\hat{x}\in I$ satisfy:
\begin{align*}
\kl{Q_{\hat{x}}}{P}&\ge \frac{b-2-1}{2m}\\
			   &= \frac{b-1}{2\frac{2(\ln(b)+2)}{(q_2-q_1)^2}}= \Omega\Bigl((q_2-q_1)^2\frac{b}{\log (b)}\Bigr),
\end{align*}
which finishes the proof of \Cref{lem:sensitivity}

\qed

\section{Discussion}

In this work we presented a limitation for the PAC-Bayes framework by showing that PAC-learnability
of one-dimensional thresholds can not be established using PAC-Bayes. 
%The result is provided in the strong distributional-independent model. In detail, we show that for any algorithm $A$, we can find a distribution $D$ where it either fails to learn or the PAC-Bayes bound in \cref{eq:thebound} is vacuous.

Perhaps the biggest caveat of our result is the mild dependence of the bound on the size of the domain in \Cref{thm:main}. In fact, \Cref{thm:main} does not exclude the possibility of PAC-learning thresholds over $\X_n$ with sample complexity that scale with $O(\log^*n)$ such that the PAC-Bayes bound vanishes. It would be interesting to explore this possibility;
one promising direction is to borrow ideas from the differential privacy literature:
\cite{Beimel16sanit} and \cite{bun2015differentially} designed a private learning algorithm for thresholds with sample complexity $\exp(\log^* n)$; this bound was later improved by \cite{kaplan2019privately} to $\tilde O((\log^* n)^2)$. 
Also, \cite{bun2020equivalence} showed that finite Littlestone dimension is sufficient for private learnability,
and it would be interesting to extend these results to the context of PAC-Bayes. 
Let us note that in the context of \emph{pure} differential privacy, 
the connection between PAC-Bayes analysis and privacy has been established in \cite{dziugaite2018data}.

\paragraph{Non-uniform learning bounds} Another aspect is the implication of our work to learning algorithms beyond the uniform PAC setting. 
Indeed, many successful and practical algorithms exhibit sample complexity that depends on the target-distribution. 
E.g.,the $k$-Nearest-Neighbor algorithm eventually learns any target-distribution (with a distribution-dependent rate).
The first point we address in this context concerns \emph{interpolating algorithms}. 
These are learners that achieve zero (or close to zero) training error (i.e.\ they interpolate the training set). 
Examples of such algorithms include kernel machines, boosting, random forests, as well as deep neural networks \cite{belkin2018overfitting, salakhotinov2017talk}. 
PAC-Bayes analysis has been utilized in this context, for example, to provide margin-dependent generalization guarantees for kernel machines~\cite{langford2003pac}. 
It is therefore natural to ask whether our lower bound has implications in this context. 
As a simple case-study, consider the $1$-Nearest-Neighbour. 
Observe that this algorithm forms a proper and consistent learner for the class of 1-dimensional thresholds\footnote{Indeed, given any realizable sample it will output the threshold which maximizes the margin.}, 
and therefore enjoys a very fast learning rate.
On the other hand, our result implies that for any algorithm (including as 1-Nearest-Neighbor) 
that is amenable to PAC-Bayes analysis, there is a distribution realizable by thresholds on which it has high population error. 
Thus, no algorithm with a PAC-Bayes generalization bound can match the performance of nearest-neighbour with respect to such distributions.

% The first observation we would like to highlight is in the context of \emph{interpolating algorithms}. Interpolating algorithms are learners that aim to achieve zero (or close to zero) training error. Such as kernel machines, boosting, random forests, \new{and it has also been suggested that the best practice for training is to first perfectly fit the data.} \cite{belkin2018overfitting, salakhotinov2017talk}. PAC-Bayes analysis has been successfully utilized, for example, in the context of large margin kernel machines, \cite{langford2003pac}. On the other hand, it is interesting to consider our lower bound result in the context of an algorithm such as Nearest-Neighbour, and the implications.

% \new{Indeed, the main observation we make is that Nearest Neighbour can easily learn the class of thresholds whenever it is given data from a realizable distribution over the real line. On the other hand, our result shows that for any Gibbs-classifier that is amenable to PAC-Bayes analysis, there must be a distribution over the real-line on which it fails to achieve zero training error or zero risk. That shows that no Gibbs-classifier with a PAC-Bayes generalization bound can match the performance of nearest-neighbour.}

Finally, this work also relates to a recent attempt to explain generalization through the implicit bias of learning algorithms: 
it is commonly argued that the generalization performance of algorithms can be explained
by an implicit algorithmic bias.
Building upon the flexibility of providing distribution-dependent generalization bounds, the PAC-Bayes framework has seen a resurgence of interest in this context towards explaining generalization in large-scale modern-time practical algorithms  \cite{neyshabur2017exploring, neyshabur2017pac, dziugaite2017computing, dziugaite2018data, arora2018stronger}. Indeed PAC-Bayes bounds seem to provide non-vacuous bounds in several relevant domains \cite{langford2002not, dziugaite2018data}. Nevertheless, the work here shows that any algorithm that can learn 1D thresholds is necessarily not biased, in the PAC-Bayes sense, 
towards a (possibly distribution-dependent) prior. We mention that recently, \cite{dauber2020can} showed that SGD's generalization performance indeed cannot be attributed to some implicit bias of the algorithm that governs the generalization.

\section*{Acknowledgements}
The authors would like to acknowledge Steve Hanneke for suggesting and encouraging them to write this manuscript.

\bibliographystyle{abbrvnat}
\bibliography{references}
\end{document}